# Homograph Disambiguation Through Selective Diacritic Restoration

**Sawsan Alqahtani,[1,2] Hanan Aldarmaki,[1] Mona Diab[1,3]**
[1]The George Washington University
[2]Princess Nourah Bint Abdul Rahman University
[3]AWS, Amazon AI
`sawsanq@gwu.edu, aldarmaki@gwu.edu, diabmona@amazon.com`

## Abstract

Lexical ambiguity, a challenging phenomenon in all natural languages, is particularly prevalent for languages with diacritics that tend to be omitted in writing, such as Arabic. Omitting diacritics leads to an increase in the number of homographs: different words with the same spelling. Diacritic restoration could theoretically help disambiguate these words, but in practice, the increase in overall sparsity leads to performance degradation in NLP applications. In this paper, we propose approaches for automatically marking a subset of words for diacritic restoration, which leads to selective homograph disambiguation. Compared to full or no diacritic restoration, these approaches yield selectively-diacritized datasets that balance sparsity and lexical disambiguation. We evaluate the various selection strategies extrinsically on several downstream applications: neural machine translation, part-of-speech tagging, and semantic textual similarity. Our experiments on Arabic show promising results, where our devised strategies on selective diacritization lead to a more balanced and consistent performance in downstream applications.

## 1 Introduction

Lexical ambiguity, an inherent phenomenon in natural languages, refers to words or phrases that can have multiple meanings. In written text, lexical ambiguity can be roughly characterized into two categories: polysemy and homonymy. A polysemous word has multiple senses that express different but related meanings (e.g. 'head' as an anatomical body part, or as a person in charge), whereas homonyms are different words that happen to have the same spelling (e.g. 'bass' as an instrument vs. a fish) (Löbner, 2013). Homographs are words that have the same spelling but may have different pronunciation and meaning.

A diacritic is a mark that is added above, below, or within letters to indicate pronunciation, vowels, or other functions. For languages that use diacritical marks, such as Arabic or Hebrew, the orthography is typically under-specified for such marks, i.e. the diacritics are omitted. This phenomenon exacerbates the lexical ambiguity problem since it increases the rate of homographs. For example, without considering context, the undiacritized Arabic word *ktb* may refer to any of the following diacritized variants:[1] *katab* كَتَب "wrote", *kutub* كُتُب "books", or *kutib* كُتِب "was written". As an illustrative analogy in English, dropping vowels in a word such as *pan* yields the underspecified token *pn* which can be mapped to *pin*, *pan*, *pun*, *pen*. It should be noted that even after fully specifying words with their relevant diacritics, homonyms such as "bass" are still ambiguous; likewise in Arabic, the fully-specified word *bayot* بَيْت can either mean "verse" or "house".

In this paper, we devise strategies to automatically identify and disambiguate a *subset* of homographs that result from omitting diacritics. While context is often sufficient for determining the meaning of ambiguous words, explicitly restoring missing diacritics should provide valuable additional information for homograph disambiguation. This process, diacritization, would render the resulting text comparable to that of languages whose words are orthographically fully specified such as English.

Past studies have focused on developing models for automatic diacritic restoration that can be used as a pre-processing step for various applications such as text-to-speech (Ungurean et al., 2008) and reading comprehension (Hermena et al., 2015). In theory, restoring all diacritics should also help improve the performance of NLP applications such as machine translation. However, in practice, full diacritic restoration results in increased sparsity and out-of-vocabulary words, which leads to degradation in performance (Diab et al., 2007; Alqahtani et al., 2016). The main objective of this work is to find a sweet spot between zero and full diacritization in order to reduce lexical ambiguity without increasing sparsity. We propose selective diacritization, a process of restoring diacritics to a subset of the words in a sentence sufficient to disambiguate homographs without significantly increasing sparsity. Selective diacritization can be viewed as a relaxed variant of word sense disambiguation since only homographs that arise from missing diacritics are disambiguated.[2]

---

[1]We adopt Buckwalter Transliteration encoding into Latin script for rendering Arabic text http://www.qamus.org/transliteration.htm.

Intrinsically evaluating the quality of a devised selective diacritization scheme against a gold set is challenging since it is difficult to obtain a dataset that exhibits consistent selective diacritization with reliable inter-annotator agreement (Zaghouani et al., 2016b; Bouamor et al., 2015), thereby necessitating an empirical automatic investigation. Hence, in this work, we evaluate the proposed selective diacritization schemes extrinsically on various semantic and syntactic downstream NLP applications: Semantic Textual Similarity (STS), Neural Machine Translation (NMT), and Part-of-Speech (POS) tagging. We compare our selective strategies against two baselines full diacritization and zero diacritics applied on all the words in the text. We use Modern Standard Arabic (MSA) as a case-study.[3]

Our approach is summarized as follows: we start with full diacritic restoration of a large corpus, then apply different unsupervised methods to identify the words that are ambiguous when undiacritized. This results in a dictionary where each word is assigned an ambiguity label (ambiguous vs. unambiguous). Selectively-diacritized datasets can then be constructed by restoring the full diacritics only to the words that are identified as ambiguous.

The contribution of this paper is threefold:

1. We introduce automatic selective diacritization as a viable step in lexical disambiguation and provide an encouraging baseline for future developments towards optimal diacritization. Section 2 describes existing work towards optimal diacritization and how they differ from our approach;

2. We propose several unsupervised data-driven methods for the automatic identification of ambiguous words;

3. We evaluate and analyze the impact of partial sense disambiguation (i.e. selective diacritic restoration of identified homographs) in downstream applications for MSA.

## 2 Related Work

We are concerned mainly with studies that target word disambiguation through the use of diacritics/accents restoration. Homograph disambiguation through accents has been explored previously in several studies with the use of different rule-based and machine-learning approaches for languages such as Arabic, Spanish, Igbo, and Vietnamese (Ezeani et al., 2017; Nguyen et al., 2012; Nivre et al., 2017; Said et al., 2013; Tufiş and Chiţu, 1999).

Bouamor et al. (2015) conducted a pilot study where they asked human annotators to add the minimum number of diacritics sufficient to disambiguate homographs. However, attempts to provide human annotation for selective diacritization resulted in low inter-annotator agreement due to the annotators' subjectivity and different linguistic understanding of the words and contexts (Bouamor et al., 2015; Zaghouani et al., 2016b). To address this issue, Zaghouani et al. (2016b) used a morphological disambiguation tool, MADAMIRA (Pasha et al., 2014), to identify candidate words that may need disambiguation. A word was considered ambiguous if MADAMIRA generates multiple high-scoring diacritic alternatives, and human annotators were asked to select from these alternatives or manually edit the diacritics if none of the options was deemed correct. This resulted in a significant increase in inter-annotator agreement. Our work differs in two aspects: first, we develop automatic methods for ambiguity detection based on word usage. We then restore the diacritics for all occurrences of these ambiguous words, whereas in (Zaghouani et al., 2016b), the same word may be tagged as ambiguous in some cases but not ambiguous in other cases depending on context, which makes it harder to generalize to new datasets.

---

[2]Identifying empirically successful selective diacritization strategies can help discover optimal diacritization schemes; however, this direction is currently beyond the scope of this work.
[3]Proposed methodologies can be applied to other languages where diacritics are omitted.

Yarowsky (1994) developed an accent restoration algorithm for Spanish and French that specifies the accent patterns for ambiguous words (i.e. multiple accent patterns). Our intuition is different than that of Yarowsky (1994) in two ways. First, they added diacritics to all words that have more than one diacritic pattern while we add the diacritics for only a subset of candidate words. Second, they used context for adding diacritics, while we use context to isolate words that require diacritics, for which we apply an off-the-shelf diacritic restoration model.

Rather than restoring all diacritics in the written text, the idea of adding diacritics sufficient to resolving lexical ambiguity was initially introduced in (Diab et al., 2007). They developed several linguistically-based partial schemes and evaluated their methods in Statistical Machine Translation. They found that fully diacritizing texts led to performance degradation due to sparseness while no diacritization increased the lexical ambiguity rate. Similar results have been found in (Alqahtani et al., 2016), where several other basic diacritic patterns were investigated. Although the impact of diacritics in machine translation was promising, the development of partial schemes does not show significant improvements over the non-diacritized and fully-diacritized baselines.

Alnefaie and Azmi (2017) introduced a partial diacritization scheme for MSA based on the output of a morphological analyzer in addition to WordNet (Black et al., 2006), and Alqahtani et al. (2018) created a lexical resource that assigned an ambiguity label for each word, where a word is considered ambiguous if it has more than one diacritic possibility, with and without considering its part-of-speech tag. However, both (Alnefaie and Azmi, 2017; Alqahtani et al., 2018) did not evaluate their methods empirically to demonstrate their effectiveness for NLP applications. Hanai and Glass (2014) similarly developed three linguistically-based partial diacritic schemes for automatic speech recognition and found statistically significant improvement over the baseline. However, their work is focused on improving word pronunciations whereas we focus on word sense disambiguation. Ezeani et al. (2017) discussed the impact of adding accents for each and every word in Igbo language, potentially increasing the performance for machine translation and word sense disambiguation.

All of the aforementioned approaches either apply full diacritics on all words whenever appropriate or derive partial diacritic schemes based on linguistic understanding; crucially these partial diacritic schemes are applied to *all* words in a sentence.[4] Our devised strategies differ in that we apply full diacritization to a *select* set of tokens in the text. Our work is related to these previous studies in the sense that we reduce the search space of candidate words that could benefit from full or partial diacritization without increasing sparsity. Furthermore, the novelty of this work lies in utilizing automatic unsupervised methods to identify such words.

## 3 Approach

### 3.1 Selective Diacritization

Selective diacritization is the process of restoring diacritics to a subset of words in a text corpus. Manually annotating words in a dataset with binary ambiguity labels (ambiguous vs. unambiguous) is challenging due to the difficulty in defining ambiguous words that would benefit from diacritics (Zaghouani et al., 2016b). Therefore, we propose several techniques to automatically identify ambiguous words for selective diacritization. Since it is common to use distributed word vector representations in downstream tasks, we define ambiguity in terms of distributional similarity among diacritized word variants. Our intuition is that variants with low distributional similarity are more likely to benefit from diacritization to disambiguate their meanings and tease apart their context variations. On the other hand, word variants with highly similar contexts tend to have very similar distributional representations, which results in unnecessary redundancy and sparsity if all variants are kept.

Based on this definition, we developed several context-based approaches to identify candidate ambiguous word types and generate a set of dictionaries with ambiguity labels (AmbigDict), where each word is marked as either ambiguous or unambiguous. The proposed approaches can be classified by the type of tokens used to create the AmbigDict: diacritized (AmbigDict-DIAC) or undiacritized (AmbigDict-UNDIAC). For example an entry in AmbigDict-UNDIAC would be "Elm" علم: ambiguous; "ktb" كتب: unambiguous, whereas in AmbigDict-DIAC would be "E**a**l**a**m" عَلَم: ambiguous; "k**utu**b" كُتُب: unambiguous.

---
[4]For instance, the undiacritized sentence *bEd ywm* بعد يوم "after a day" would be diacritized as *baEod yawom* بَعْد يَوْم when fully diacritized, ***bEod ywom*** بَعْد يَوْم ((Diab et al., 2007; Alqahtani et al., 2016)'s SUK scheme) when partially diacritized, ***baEod*** *ywm* بَعْد يوم when selectively diacritized.

## 3.2 AmbigDict-UNDIAC Generation

We explore two methods for creating ambiguity dictionaries from undiacritized text: using a morphological analyzer, and unsupervised sense induction.

**Multiple Morphological Variants (MULTI):** The number of diacritic alternatives for a word can be a clue to determine whether a word is ambiguous due to missing diacritics (Alqahtani et al., 2018). In this approach, context is not considered, but rather the output of a morphological analyzer applied to the text. We leverage the morphological analyzer component of MADAMIRA (Pasha et al., 2014) to generate all possible valid diacritic variants of a word whether these variants are present in the corpus or not. If an undiacritized word has more than one possible diacritic variant, we consider it ambiguous. We use this context-independent approach as a baseline.

**Sense Induction Based Approach (SENSE):** Selective diacritization is related to word sense disambiguation, however we only target disambiguation through diacritic restoration. Techniques used in automatic word sense induction can be used as a basis for identifying ambiguous words. Using undiacritized text, we apply an off-the-shelf system for word sense induction developed by Pelevina et al. (2017), which uses the Chinese Whispers algorithm (Biemann, 2006) to identify senses of a graph constructed by computing the word similarities (highest cosine similarities) through using word as well as context embeddings. We apply the first three steps described in Pelevina et al. (2017) but we do not use the generated sense-based embeddings; we only use the system to identify the words with multiplw senses. We set the three parameters as follows: the graph size $N$ to 200, the inventory granularity $n$ to 400, and the minimum number of clusters (senses) $k$ to 5.[5] A word type is deemed ambiguous if it appears in more than one cluster.

## 3.3 AmbigDict-DIAC Generation

We explore clustering and translation based methods to create ambiguity dictionaries from diacritized text.

**Clustering-based Approaches (CL):** Similar in spirit to SENSE, we apply unsupervised clustering to our corpora to induce AmbigDict. However, unlike SENSE, we apply clustering to diacritized data. Our intuition is that dissimilar words are likely to occur in different contexts, and therefore likely to be in different clusters. Therefore, we tag words as ambiguous if diacritized variants of the same underlying undiacritized form appear in different clusters.

As a preprocessing step, we apply a full contextualized diacritization tool to the underlying corpora. We leverage the MADAMIRA tool (Pasha et al., 2014) to produce fully diacritized text (for every token in the data) covering both types of diacritic restoration: lexical and syntactic. The latter covers syntactic case and mood diacritics. In this study, we are only concerned with lexical ambiguity; Moreover, MADAMIRA has a very high diacritic error rate in syntactic diacritic restoration (15%) compared to (3.5%) for lexical diacritic restoration. Hence, we drop the predicted word final syntactic diacritics resulting in a diacritization scheme similar to the partial scheme in (Diab et al., 2007; Alqahtani et al., 2016), namely, FULL-CM. In FULL-CM, every token is fully lexically diacritized (e.g. the fully diacritized words Ealam**a** عَلَمَ and Ealam**u** عَلَمُ differ in their syntactic diacritics and are mapped to Ealam عَلَم "flag" in FULL-CM).

Given this diacritized corpus, we apply three different standard clustering approaches: Brown[6] (Brown et al., 1992) (CL-BR), K-means[7] (Kanungo et al., 2002) (CL-KM), and Gaussian Mixture via Expectation Maximization (CL-EM)[8] (Dempster et al., 1977). We tune the number of clusters for downstream tasks; in particular, we empirically investigate the performance on the de-

---

[5] We tuned these parameters empirically.

[6] https://github.com/percyliang/brown-cluster

[7] We use "sickit-learn" version 0.18.1. We use the value 1 for both random_state and n_init and the default values for the remaining parameters.

[8] We use "sickit-learn" version 0.18.1. with the following parameters: max_iter=1000, random_state=1, and covariance_type=spherical

velopment set in the downstream tasks for different number of clusters.

**Translation-based Approaches (TR):** Translation can be used as a basis for word sense induction (Diab and Resnik, 2002; Ng et al., 2003) since words across different languages tend to have disparate senses. Following a similar intuition, we use English translations from a parallel corpus as a trigger to divide the set of diacritic possibilities of a word into multiple subsets. The intuition here is that homographs worth disambiguating are those that are likely to be translated differently. We leverage an English MSA parallel corpus, where the MSA is diacritized in the Full-CM scheme using MADAMIRA (the same preprocessing step for CL described above). In this approach, diacritized variants that share the same English translations are considered unambiguous, whereas those that are typically translated to different English words are considered ambiguous. To that end, we first align the sentences at the token level and generate word translation probabilities using fast-align (Dyer et al., 2013), which is a log-linear reparameterization of IBM Model 2 (Brown et al., 1993). If a word shares any translation with its diacritized variant in the top $N$ most likely translations, we consider it unambiguous (e.g. Ealam عَلَم "flag" and Ealima عَلِمَ "learned" are unambiguous since they do not share top translations). Otherwise, the word is tagged as ambiguous. We tune $N$ to include 1, 5, 10, and all translations.

## 4 Evaluation

Once we have generated the two variants of AmbigDict (AmbigDict-UNDIAC and AmbigDict-DIAC), we evaluate their efficacy extrinsically on downstream applications. For all downstream applications, training and test data are preprocessed using MADAMIRA (Pasha et al., 2014) with the FULL-CM diacritization scheme where we only keep lexical diacritics.[9] Then the data is filtered based on the AmbigDict of choice; namely, only word tokens in the text deemed ambiguous according to each AmbigDict maintain their full diacritics (as generated by MADAMIRA) while the unambiguous words are kept undiacritized.

---

[9] Full diacritics are included except inflectional diacritics that reflect the syntactic positions of words within sentences but do not alter meaning.

### 4.1 Datasets

For MULTI, SENSE, CR, we use a combination of four Modern Standard Arabic (MSA) datasets that vary in genre and domain and add up to ∼50M tokens: Gigaword 5th edition, distributed by Linguistic Data Consortium (LDC), Wikipedia dump 2016, Corpus of Contemporary Arabic (CCA) (Zaghouani et al., 2016a; Al-Sulaiti and Atwell, 2006), and LDC Arabic Tree Bank (ATB).[10] For TR, we use an Arabic-English parallel dataset which includes ∼60M tokens and is created from 53 LDC catalogs. For data cleaning, we replace e-mails and URLs with a unified token and use SPLIT tool (Al-Badrashiny et al., 2016) to clean UTF8 characters (e.g. Latin and Chinese), remove diacritics in the original data, and separate punctuation, symbols, and numbers in the text, and replace them with separate unified tokens. We split long sentences (more than 150 words) by punctuation and then remove all sentences that are still longer than 150 words. We use D3 style (i.e. all affixes are separated) (Pasha et al., 2014) for Arabic tokenization without normalizing characters. For English, we lower case all characters and use TreeTagger (Schmid, 1999) for tokenization. We used SkipGram word embeddings (Mikolov et al., 2013), where applicable.

### 4.2 Extrinsic Evaluation

We evaluate the effectiveness of the proposed approaches using three applications: Semantic Textual Similarity (STS), Neural Machine Translation (NMT), and Part-of-Speech (POS) tagging. We used different significance testing methods appropriate for each application with $p = 0.05$.

#### 4.2.1 Semantic Textual Similarity (STS)

STS is a benchmark evaluation task (Cer et al., 2017), where the objective is to predict the similarity score between a pair of sentences. Performance is typically evaluated using the Pearson correlation coefficient against human judgments. We used the William test (Graham and Baldwin, 2014) for significance testing. We experiment with an unsupervised system based on matrix factorization developed by (Guo and Diab, 2012; Guo et al., 2014), which generates sentence embeddings from a word-sentence co-occurrence matrix, then compare them using cosine similarity. We use a dimension size of 700. To train the model, we use the

---

[10] Parts 1, 2, 3, 5, 6, 7, 10, 11, and 12

Arabic dataset released for SemEval-2017 task 1 (Cer et al., 2017). Since the training dataset is small, we augment it by randomly selecting sentences from the dataset (∼1,655,922) described in Section 4.1 where the chosen sentences have to satisfy the following conditions: the number of words lie between 5 and 150; and, the minimum frequency for each word is 2. We apply these conditions in the diacritized data since it suffers more from sparseness, and then use their undiacritized correspondents in the undiacritized setting.

### 4.2.2 Neural Machine Translation (NMT)

We build a BiLSTM-LSTM encoder-decoder machine translation system as described in (Bahdanau et al., 2014) using OpenNMT (Klein et al., 2014). We use 300 as input dimension for both source and target vectors, 500 as hidden units, and 0.3 for dropout. We initialize words with embeddings trained using FastText (Bojanowski et al., 2017) on the selectively-diacritized dataset described in Section 4.1. We train the model using SGD with max gradient norm of 1 and learning rate decay of 0.5. We use the Web Inventory of Transcribed and Translated Talks (WIT), which is made available for IWSLT 2016 (Mauro et al., 2012). We use BLEU (Papineni et al., 2002) for evaluation, and bootstrap re-sampling and approximate randomization for significance testing (Clark et al., 2011).

### 4.2.3 POS tagging

POS tagging is the task of determining the syntactic role of a word (i.e. part of speech) within a sentence. We use a BiLSTM-CRF architecture to train a POS tagger using the implementation provided by (Reimers and Gurevych, 2017), with 300 as dimension size, initialized using the same embeddings we use in NMT. We used ATB datasets parts 1,2, and 3 to train the models with Universal Dependencies POS tags, version 2 (Nivre et al., 2016). We use word-level accuracy for evaluation, and t-test (Fisher, 1935; Dror et al., 2018) for significance testing.

### 4.3 Automatic Diacritization

For generating the various AmbigDict approaches, we used either fully diacritized versions, without case and mood related diacritics,[11] or undiacritized versions of the datasets. Since it is expensive to obtain enormous human-annotated diacritized datasets, we use the morphological analysis and disambiguation tool, MADAMIRA version 2016 2.1 (Pasha et al., 2014)

### 4.4 AmbigDict Statistics

Table 1 shows the number of identified ambiguous words using each approach. Note that the total vocabulary sizes vary due to either different datasets (e.g. for TR) or different preprocessing (e.g. MULTI is based on undiacritized text). For a given corpus, the number of ambiguous words identified by MULTI can be viewed as an estimate of the upper bound on ambiguous words due to diacritics. In MULTI, words that have no valid analysis generated by MADAMIRA are filtered; this resulted in significant drop of the number of types since the dataset includes noisy and infrequent instances.

| Dictionary | Types | % Ambig Words |
|---|---|---|
| **AmbigDict-UNDIAC** | | |
| MULTI | 168,384 | 33.82 |
| SENSE | 467,953 | 8.50 |
| **AmbigDict-DIAC** | | |
| CL | 497,222 | 8.70 - 8.98 |
| TR | 36,533 | 27.58 |

Table 1: Vocabulary size and percentage of ambiguous entries in AmbigDict-DIAC and AmbigDict-UNDIAC.

### 4.5 Results and Analysis

| Dictionary | STS | NMT | POS |
|---|---|---|---|
| NONE | **0.608** | **27.1** | 97.99% |
| FULL-CM | 0.593 | 26.8 | **98.06%** |
| **AmbigDict-UNDIAC** | | | |
| MULTI | 0.591 | 27.0 | **98.11%**\* |
| SENSE | 0.598 | 27.1 | 97.97% |
| **AmbigDict-DIAC** | | | |
| CL-BR | 0.601 | 27.1 | **98.09%** |
| CL-KM | 0.608 | **27.2** | 98.05% |
| CL-EM | **0.617**\* | 27.1 | 98.05% |
| TR | **0.616**\* | **27.3**\* | 97.94% |

Table 2: Performance with selectively-diacritized datasets in downstream applications. **Bold** numbers indicate approaches with higher performance than the best performing baseline. \* refers to approaches with statistically-significant performance gains against the best performing baseline.

Table 2 shows the performance of all strategies in downstream tasks. Comparing baselines NONE

---
[11] FULL-CM diacritization scheme, where we only keep lexical diacritics.

and FULL-CM, we observe that applications that require semantic understanding (STS and NMT) show better performance when the dataset is undiacritized, whereas POS tagging yields better performance with the fully diacritized dataset.

The differences in performance between the baselines are significant across all tasks. In all tasks, at least one of the selective diacritization schemes leads to performance gains compared to both baselines. However, the choice of best performing selective diacritization scheme varies across tasks. In general, AmbigDict-DIAC approaches provide more promising results on semantic related applications.

TR and CL-EM approaches yield the highest performance in two of the applications (STS and NMT), while MULTI and CL-BR achieved the highest performance in POS tagging. Incidentally, MULTI has the highest rate of ambiguous words, which leads to more disambiguation through diacritization. This is consistent with the observation that diacritization is useful for syntactic tasks like POS tagging, as observed through the baselines. In all other tasks, all selective diacritization schemes performed significantly higher than full diacritization.

**Homograph Evaluation:** We compared the performance of the various schemes on subsets of the test sets that include homographs, which are identified from the FULL-CM version of the training datasets. For STS and NMT evaluation, we kept only the test sentences that contain at least one homograph. For POS word-level evaluation, we only considered the homographs. Table 3 shows homograph performance across applications. The performance on these subsets follow the same trend as the overall results illustrated in Table 2 except for POS tagging, where FULL-CM achieved the comparable performance to the selective schemes. Note, however, that almost all schemes achieved higher POS tagging accuracy than NONE in these subsets, and almost all schemes achieved comparable or higher performance than FULL-CM in STS and NMT, with TR significantly outperforming the rest of the schemes as well as the baselines. This illustrates the usefulness of selective diacritization for balancing homograph disambiguation and sparsity compared to full or no diacritization.

**Frequent POS Tag Performance:** POS tagging labels each word in the sentence as opposed to NMT and STS which are evaluated at the sentence level. Thus, we compared the best performing scheme (MULTI) and the baselines in terms of their per tag performance on the four most frequent tags: verbs, nouns, adjectives, and adverbs. Table 4 shows the results of the baselines and MULTI. For verbs and nouns, MULTI has better performance than both baselines followed by FULL-CM. For adjectives and adverbs, NONE followed by MULTI have better performance than FULL-CM. While FULL-CM has overall higher accuracy, these results indicate that selective diacritization is a better approach for the most frequent tags, possibly due to reduced sparsity compared with FULL-CM.

| Dictionary | STS | NMT | POS |
|---|---|---|---|
| NONE | **0.590** | **27.4** | 98.26% |
| FULL-CM | 0.575 | 27 | **98.70%** |
| AmbigDict-UNDIAC | | | |
| MULTI | 0.574 | 27.2 | 98.65% |
| SENSE | 0.581 | 27.3 | 98.37% |
| AmbigDict-DIAC | | | |
| CL-BR | 0.584 | 27.4 | 98.59% |
| CL-KM | **0.591** | **27.5** | 98.52% |
| CL-EM | **0.60**\* | **27.4** | 98.47% |
| TR | **0.597**\* | **27.6**\* | 98.22% |

Table 3: Performance of selectively-diacritized datasets on homographs. **Bold** numbers indicate approaches with higher performance than the best performing baseline. \* refers to approaches with statistically-significant performance gains against the best performing baseline.

**OOV Performance:** We measured the POS tagging performance on Out-of-Vocabulary (OOV) words to measure the effect of sparsity on performance. We consider a word OOV if it does not occur in the fully-diacritized training set. FULL-CM achieved 87.43% tag accuracy, while NONE achieved 87.56%. Using the MULTI scheme, the POS tagging accuracy on OOV words was 87.51%, which falls between the two baselines, as expected.

The results above indicate that using a selective diacritization scheme like MULTI can achieve a desirable balance between disambiguation and sparsity, such that better performance can be achieved in the frequent cases without increasing sparsity and OOV rates.

| Scheme | Verb | Noun | Adj | Adv |
|---|---|---|---|---|
| MULTI | **95.98%** | **97.63%** | 94.43% | 97.05% |
| NONE | 95.08% | 97.45% | **94.71%** | **98.08%** |
| FULL-CM | 95.87% | 97.56% | 94.40% | 96.79% |

Table 4: POS Tagging performance per most frequent tag. **Bold** scores indicate the highest score in a column.

| Type | Example |
|---|---|
| part-of-speech | $ak شَك "doubt" (noun) |
| | $ak~ شَكّ "doubted" (verb) |
| action direction | >a*okur أَذْكُر "remember" |
| | >u*ak~ir أُذَكِّر "remind" |
| number | $uyuwEiy~ayon شُيُوعِيَّيْن "communists" |
| | $uyuwEiy~iyn شُيُوعِيِّين "communists" |

Table 5: Examples of ambiguous word pairs detected by the clustering approaches.

### 4.6 Properties of Ambiguity Dictionaries

**Clustering-Based Ambiguity:** While MULTI, TR, and SENSE approaches have intuitive justifications, the clustering approaches are based entirely on distributional features. We analyzed some of the results qualitatively to shed light on types of words that are deemed ambiguous through clustering. While the various clustering approaches resulted in different labeling, their overall statistics and patterns were highly similar. Using a random subset of words from these CL dictionaries, we extracted the examples shown in Table 5, which shows some of the most common types of ambiguity. Note that the detected words are either semantically ambiguous (e.g. derivations or distinct lemmas) or syntactically ambiguous (e.g. part-of-speech).

**Diacritic Pattern Complexity:** We investigated whether there are regular diacritic patterns among words that were considered ambiguous by CL and TR. Both approaches are data-driven, and we applied them on different corpora, so we investigated their degree of agreement. To do so, we abstracted the diacritic patterns for words in the vocabulary by converting all characters other than diacritics to a unified token "C", then we collected statistics of patterns of word pairs that are deemed ambiguous vs. unambiguous. For example, the ambiguous pair "katab" كَتَب and "kutib" كُتِب have the pattern CaCaC-CuCiC. For CL methods, the number of unique diacritic patterns of unambigu-ous word pairs (i.e. falling in the same cluster) were between 197-219 patterns, whereas patterns of ambiguous pairs were between 813-872. The majority of patterns between unambiguous words also occurred between ambiguous words. For TR, while most patterns were labeled unambiguous, around 300 patterns were always labeled ambiguous. We did not find overarching semantic or syntactic rules that consistently explain ambiguity tags. However, a number of patterns ($\sim 20$) were always tagged as ambiguous by both TR and CL approaches. Table 6 shows a sample of these patterns with examples.

| Pattern Pair | Example |
|---|---|
| CaC~aC | Ear~aD عَرَّض "make wider" |
| CuCiC | EuriD عُرِض "has been shown" |
| CaCiCaCoC | ba$iEayon بَشِعَيْن "ugly" (dual) |
| CaCiCiCC | ba$iEiyn بَشِعِين "ugly" (plural) |

Table 6: Examples of consistent diacritic patterns of ambiguous words between CL and TR approaches.

## 5 Discussion & Conclusion

We investigated selective diacritization as a viable technique for reducing lexical ambiguity using Arabic as a case study. To our knowledge, this is the first work that shows encouraging results with automatic selective diacritization schemes in which the devised approaches evaluated on several downstream applications. Our findings demonstrate that partial diacritization achieves a balance between homograph disambiguation and sparsity effects; the performance using selective diacritization always approached the best of both extremes in each application, and sometimes surpassed the performance of both baselines, which is consistent with our intuition of balancing sparsity and disambiguation for improving overall performance.

While the increase in performance was not consistent across all tasks, the results provide an empirical evidence of the viability of automatic partial diacritization, especially since manual efforts in this vein had been rather challenging. We believe that the approaches described in this paper could help advance the efforts towards optimal diacritization schemes, which are currently mostly based on linguistic features. We analyzed some patterns that were recognized as ambiguous using our best-performing schemes, and showed some consistencies in the diacritic patterns, although the results were not conclusive. We believe that a deeper analysis of these patterns may help shed light on the lexical ambiguity phenomenon in addition to allowing further improvements in selective diacritization.

# References

Mohamed Al-Badrashiny, Arfath Pasha, Mona Diab, Nizar Habash, Owen Rambow, Wael Salloum, and Ramy Eskander. 2016. SPLIT: Smart preprocessing (Quasi) language independent tool. In *International Conference on Language Resources and Evaluation (LREC)*.

Latifa Al-Sulaiti and Eric Atwell. 2006. The design of a corpus of contemporary Arabic. *International Journal of Corpus Linguistics*, 11(2):135–171.

Rehab Alnefaie and Aqil M. Azmi. 2017. Automatic minimal diacritization of Arabic texts. In *3rd International Conference on Arabic Computational Linguistics (ACLing)*.

Sawsan Alqahtani, Mona Diab, and Wajdi Zaghouani. 2018. ARLEX: A large scale comprehensive lexical inventory for Modern Standard Arabic. In *OSACT 3: The 3rd Workshop on Open-Source Arabic Corpora and Processing Tools*.

Sawsan Alqahtani, Mahmoud Ghoneim, and Mona Diab. 2016. Investigating the impact of various partial diacritization schemes on Arabic-English statistical machine translation. In *International Association for Machine Translation in the Americas (AMTA)*.

Dzmitry Bahdanau, Kyunghyun Cho, and Yoshua Bengio. 2014. Neural machine translation by jointly learning to align and translate. In *International Conference on Learning Representations*.

Chris Biemann. 2006. Chinese whispers: an efficient graph clustering algorithm and its application to natural language processing problems. In *Proceedings of the first workshop on graph based methods for natural language processing*. Association for Computational Linguistics.

William Black, Sabri Elkateb, Horacio Rodriguez, Musa Alkhalifa, Piek Vossen, Adam Pease, and Christiane Fellbaum. 2006. Introducing the Arabic WordNet project. In *Proceedings of the third international WordNet conference*.

Piotr Bojanowski, Edouard Grave, Armand Joulin, and Tomas Mikolov. 2017. Enriching word vectors with subword information. *Transactions of the Association for Computational Linguistics*.

Houda Bouamor, Wajdi Zaghouani, Mona Diab, Ossama Obeid, Kemal Oflazer, Mahmoud Ghoneim, and Abdelati Hawwari. 2015. A pilot study on Arabic multi-genre corpus diacritization. In *Proceedings of the Second Workshop on Arabic Natural Language Processing*.

Peter F Brown, Peter V Desouza, Robert L Mercer, Vincent J Della Pietra, and Jenifer C Lai. 1992. Class-based n-gram models of natural language. *Computational linguistics*, 18(4):467–479.

Peter F Brown, Vincent J Della Pietra, Stephen A Della Pietra, and Robert L Mercer. 1993. The mathematics of statistical machine translation: parameter estimation. *Computational linguistics*, 19(2):263–311.

Daniel Cer, Mona Diab, Eneko Agirre, Inigo Lopez-Gazpio, and Lucia Specia. 2017. Semeval-2017 task 1: Semantic textual similarity-multilingual and cross-lingual focused evaluation. In *SemEval workshop at ACL*.

Jonathan H Clark, Chris Dyer, Alon Lavie, and Noah A Smith. 2011. Better hypothesis testing for statistical machine translation: Controlling for optimizer instability. In *Proceedings of the 49th Annual Meeting of the Association for Computational Linguistics: Human Language Technologies*, pages 176–181.

Arthur P Dempster, Nan M Laird, and Donald B Rubin. 1977. Maximum likelihood from incomplete data via the EM algorithm. *Journal of the royal statistical society.*, 39:1–38.

Mona Diab, Mahmoud Ghoneim, and Nizar Habash. 2007. Arabic diacritization in the context of statistical machine translation. In *Proceedings of MT-Summit*.

Mona Diab and Philip Resnik. 2002. An unsupervised method for word sense tagging using parallel corpora. In *Proceedings of the 40th Annual Meeting on Association for Computational Linguistics*, pages 255–262.

Rotem Dror, Gili Baumer, Segev Shlomov, and Roi Reichart. 2018. The hitchhikers guide to testing statistical significance in natural language processing. In *Proceedings of the 56th Annual Meeting of the Association for Computational Linguistics*, volume 1, pages 1383–1392.

Chris Dyer, Victor Chahuneau, and Noah A Smith. 2013. A simple, fast, and effective reparameterization of ibm model 2. In *Proceedings of the 2013 Conference of the North American Chapter of the Association for Computational Linguistics: Human Language Technologies*, pages 644–648.

Ignatius Ezeani, Mark Hepple, and Ikechukwu Onyenwe. 2017. Lexical disambiguation of Igbo using diacritic restoration. In *Proceedings of the 1st Workshop on Sense, Concept and Entity Representations and their Applications*, pages 53–60.

Ronald Aylmer Fisher. 1935. *The design of experiments.* Oliver And Boyd.

Yvette Graham and Timothy Baldwin. 2014. Testing for significance of increased correlation with human judgment. In *Proceedings of the 2014 Conference on Empirical Methods in Natural Language Processing (EMNLP)*, pages 172–176.

Weiwei Guo and Mona Diab. 2012. Modeling sentences in the latent space. In *Proceedings of the 50th Annual Meeting of the Association for Computational Linguistics*, pages 864–872.

Weiwei Guo, Wei Liu, and Mona Diab. 2014. Fast tweet retrieval with compact binary codes. In *Proceedings of COLING 2014, the 25th International Conference on Computational Linguistics: Technical Papers*, pages 486–496.

Tuka Al Hanai and James R Glass. 2014. Lexical modeling for Arabic ASR: A systematic approach. In *Fifteenth Annual Conference of the International Speech Communication Association*.

Ehab W Hermena, Denis Drieghe, Sam Hellmuth, and Simon P Liversedge. 2015. Processing of Arabic diacritical marks: Phonological–syntactic disambiguation of homographic verbs and visual crowding effects. *Journal of Experimental Psychology: Human Perception and Performance*, 41(2).

Tapas Kanungo, David M Mount, Nathan S Netanyahu, Christine D Piatko, Ruth Silverman, and Angela Y Wu. 2002. An efficient k-means clustering algorithm: Analysis and implementation. *IEEE Transactions on Pattern Analysis & Machine Intelligence*, pages 881–892.

G. Klein, Y. Kim, Y. Deng, J. Senellart, and A. M. Rush. 2014. OpenNMT: Open-Source Toolkit for Neural Machine Translation. In *Proceedings of ACL 2017, System Demonstrations*.

Sebastian Löbner. 2013. *Understanding semantics*. Routledge.

Cettolo Mauro, Girardi Christian, and Federico Marcello. 2012. Wit3: Web inventory of transcribed and translated talks. In *Conference of European Association for Machine Translation*, pages 261–268.

Tomas Mikolov, Kai Chen, Greg Corrado, and Jeffrey Dean. 2013. Efficient estimation of word representations in vector space.

Hwee Tou Ng, Bin Wang, and Yee Seng Chan. 2003. Exploiting parallel texts for word sense disambiguation: an empirical study. In *Proceedings of the 41st Annual Meeting on Association for Computational Linguistics*, pages 455–462.

Minh Trung Nguyen, Quoc Nhan Nguyen, and Hong Phuong Nguyen. 2012. Vietnamese diacritics restoration as sequential tagging. In *IEEE RIVF International Conference on Computing Communication Technologies, Research, Innovation, and Vision for the Future*, pages 1–6.

Joakim Nivre, Marie-Catherine De Marneffe, Filip Ginter, Yoav Goldberg, Jan Hajic, Christopher D Manning, Ryan McDonald, Slav Petrov, Sampo Pyysalo, Natalia Silveira, Reut Tsarfaty, and Daniel Zeman. 2017. A Bambara tonalization system for word sense disambiguation using differential coding, segmentation and edit operation filtering. In *Proceedings of the Eighth International Joint Conference on Natural Language Processing*, volume 1, pages 694–703.

Joakim Nivre, Marie-Catherine De Marneffe, Filip Ginter, Yoav Goldberg, Jan Hajic, Christopher D Manning, Ryan T McDonald, Slav Petrov, Sampo Pyysalo, Natalia Silveira, et al. 2016. Universal dependencies v1: A multilingual treebank collection. In *Proceedings of the Tenth International Conference on Language Resources and Evaluation (LREC)*.

Kishore Papineni, Salim Roukos, Todd Ward, and Wei-Jing Zhu. 2002. BLEU: a method for automatic evaluation of machine translation. In *Proceedings of the 40th annual meeting on association for computational linguistics*, pages 311–318.

Arfath Pasha, Mohamed Al-Badrashiny, Mona T Diab, Ahmed El Kholy, Ramy Eskander, Nizar Habash, Manoj Pooleery, Owen Rambow, and Ryan Roth. 2014. MADAMIRA: A fast, comprehensive tool for morphological analysis and disambiguation of Arabic. In *LREC*, volume 14, pages 1094–1101.

Maria Pelevina, Nikolay Arefyev, Chris Biemann, and Alexander Panchenko. 2017. Making sense of word embeddings. In *Proceedings of the 1st Workshop on Representation Learning for NLP on Association for Computational Linguistics*.

Nils Reimers and Iryna Gurevych. 2017. Reporting score distributions makes a difference: performance study of LSTM-networks for sequence tagging. In *Proceedings of the 2017 Conference on Empirical Methods in Natural Language Processing (EMNLP)*, pages 338–348.

Ahmed Said, Mohamed El-Sharqwi, Achraf Chalabi, and Eslam Kamal. 2013. A hybrid approach for Arabic diacritization. In *International Conference on Application of Natural Language to Information Systems*, pages 53–64.

Helmut Schmid. 1999. Improvements in part-of-speech tagging with an application to German. In *Natural Language Processing Using Very Large Corpora*, pages 13–25.

Dan Tufiş and Adrian Chiţu. 1999. Automatic diacritics insertion in Romanian texts. In *Proceedings of the International Conference on Computational Lexicography COMPLEX*, volume 99, pages 185–194.

Cătălin Ungurean, Dragoş Burileanu, Vladimir Popescu, Cristian Negrescu, and Aurelian Dervis.

2008. Automatic diacritic restoration for a TTS-based e-mail reader application. *UPB Scientific Bulletin, Series C*, 70(4):3–12.

David Yarowsky. 1994. Decision lists for lexical ambiguity resolution: application to accent restoration in Spanish and French. In *Proceedings of the 32nd annual meeting on Association for Computational Linguistics*, pages 88–95.

Wajdi Zaghouani, Houda Bouamor, Abdelati Hawwari, Mona T Diab, Ossama Obeid, Mahmoud Ghoneim, Sawsan Alqahtani, and Kemal Oflazer. 2016a. Guidelines and framework for a large scale Arabic diacritized corpus. In *The Tenth International Conference on Language Resources and Evaluation (LREC)*, page 36373643.

Wajdi Zaghouani, Abdelati Hawwari, Sawsan Alqahtani, Houda Bouamor, Mahmoud Ghoneim, Mona Diab, and Kemal Oflazer. 2016b. Using ambiguity detection to streamline linguistic annotation. In *Proceedings of the Workshop on Computational Linguistics for Linguistic Complexity (CL4LC)*, pages 127–136.